\documentclass[runningheads]{llncs}

\usepackage{eccv}

\usepackage{eccvabbrv}

\usepackage{graphicx}
\usepackage{booktabs, tabularx}

\usepackage[accsupp]{axessibility}  %

\usepackage{nicematrix}

\usepackage{amsmath}
\usepackage{booktabs}
\usepackage{array,multirow,graphicx}
\usepackage{floatrow}
\usepackage{algorithm}
\usepackage{algpseudocode}
\usepackage{subcaption}
\usepackage{nicefrac}
\usepackage{eqnarray}

\DeclareMathOperator*{\argmin}{arg\,min}
\newcommand{\norm}[1]{\left\lVert#1\right\rVert}

\usepackage{hyperref}

\usepackage{orcidlink}

\begin{document}

\title{Deep Patch Visual SLAM}

\author{Lahav Lipson \and
Zachary Teed \and
Jia Deng}

\authorrunning{Lipson, Teed, Deng}

\institute{Princeton University}

\maketitle

\begin{abstract}
Recent work in visual SLAM has shown the effectiveness of using deep network backbones. Despite excellent accuracy, however, such approaches are often expensive to run or do not generalize well zero-shot. Their runtime can also fluctuate wildly while their frontend and backend fight for access to GPU resources. To address these problems, we introduce Deep Patch Visual (DPV) SLAM, a method for monocular visual SLAM on a single GPU. DPV-SLAM maintains a high minimum framerate and small memory overhead (5-7G) compared to existing deep SLAM systems. On real-world datasets, DPV-SLAM runs at 1x-4x real-time framerates. We achieve comparable accuracy to DROID-SLAM on EuRoC and TartanAir while running 2.5x faster using a fraction of the memory. DPV-SLAM is an extension to the DPVO visual odometry system; its code can be found in the same repository: \url{https://github.com/princeton-vl/DPVO}
\end{abstract}

\section{Introduction}
\label{sec:intro}

Visual Simultaneous Localization and Mapping (SLAM) is an online variant of Structure-from-Motion, where the input is a video stream. SLAM is a longstanding problem from robotics, where planning and control algorithms rely on accurate real-time state estimation. Recently, visual SLAM has also been utilized as a sub-system for computer vision algorithms, including those for monocular depth~\cite{yin2023metric3d, koltundepth}, view synthesis~\cite{rosinol2023nerf, goslam}, and 3D human pose~\cite{yin2024whac, shin2024wham}. In particular, the camera pose predictions from visual SLAM enable one to easily ground multiple single-image predictions in a global reference frame. The camera poses also enable 3D reconstruction and view-synthesis methods to leverage constraints from epipolar geometry.

A key challenge of these settings is that they often deal with monocular video with no inertial measurements, requiring a SLAM solution with higher tracking accuracy. Therefore, several computer vision algorithms~\cite{rosinol2023nerf, shin2024wham, ye2023decoupling, yin2023metric3d, deflowslam, ye2023pvo} are built-upon or utilize deep-network-based SLAM algorithms for providing camera pose, which are typically more accurate than their classical counterparts. However, such algorithms~\cite{deflowslam, goslam, teed2021droid} require a GPU with 24G of memory to run on all standard datasets, making them viable for offline pre-processing but less suitable for online applications.\looseness=-1%

To address this high cost, Deep Patch Visual Odometry~\cite{teed2023deep} (DPVO) was introduced, which borrowed the same overall mechanism as DROID-SLAM~\cite{teed2021droid}, but removed the requirement for dense correspondence in favor of sparse optical flow, making the system significantly cheaper for camera pose estimation. However, DPVO is only an odometry system, meaning it contains no mechanism to correct for accumulated pose errors. A fully-fledged SLAM system will typically detect previously visited locations and perform global optimization at a low frequency and in a separate thread (i.e., a backend).

Unfortunately, this odometry+backend paradigm fundamentally breaks for SLAM systems based on deep networks, since two separate CUDA operations on the same device will usually run sequentially, not in parallel, even when called from separate processes. The consequence is that, on a single GPU, existing deep SLAM systems will periodically drop from $\approx$30hz to <1hz while waiting for a single iteration of their backend to run. Consistent real-time inference thus requires two separate GPUs.

A secondary challenge for deep SLAM systems is that global optimization requires storing a significant number of deep features in GPU memory. For approaches based on optical flow~\cite{deflowslam, teed2021droid, goslam}, this means retaining expensive dense feature maps for all frames in anticipation that they may be used by the backend at some point, leading to GPU memory usage which grows quickly with the video length.

In this work, we extend the DPVO odometry system to a full SLAM system by introducing a mechanism for loop closure which does not suffer from these issues, and refer to the full system as ``DPV-SLAM''. We also introduce a separate, traditional loop closure mechanism based on classical features, with benefits complementary to typical deep-SLAM backends. This is primarily a systems paper; we aim to provide an efficient and accurate utility for camera pose estimation for in-the-wild video with global optimization and loop closure. We describe our contributions in detail and open source our code. We hope this will be a valuable resource for the community. We evaluate our approach on EuRoC, KITTI, TUM-RGBD and TartanAir, and observe:

\begin{itemize}
    \item \textbf{DPV-SLAM is accurate.} We perform similarly to DROID-SLAM on EuRoC and TartanAir. Compared to DPVO, we achieve 4x lower error on EuRoC (0.105$\rightarrow$0.024). On KITTI, we outperform DROID-SLAM and DPVO by significant margins.
    \item \textbf{DPV-SLAM is fast and efficient.} DPV-SLAM runs 2.5x faster than DROID-SLAM on EuRoC and 2.3x faster on KITTI. Compared to the base DPVO system, we incur only a small reduction in speed (e.g., 60$\rightarrow$50 FPS on \cite{burri2016euroc}) and increase in cost (4G$\rightarrow$5G GPU memory).
    \item \textbf{DPV-SLAM is general and robust.} Our method performs well in many settings, suffering from 0 catastrophic failures. We compare DPV-SLAM to other methods which report results on both indoor/outdoor without re-training, and show that our method does not struggle in any domain.
\end{itemize}
To construct DPV-SLAM, we introduce two efficient mechanisms to correct drift: a \emph{proximity} loop closure and a \emph{classical} one. The former uses camera proximity to detect loops, and addresses a challenge with building SLAM systems on deep networks, which is their inability to run the backend and frontend in parallel. It doesn't require a separate GPU, and is inexpensive and fast to run. The central idea is to optimize a single, shared scene graph with both odometry and low-cost loop closure factors mixed together. To enable efficient global optimization, we contribute a CUDA-accelerated block-sparse implementation of bundle adjustment which is compatible with DPVO's ``patch graph'' scene representation. Our proximity-based loop closure runs considerably faster DROID-SLAM's backend on EuRoC~\cite{burri2016euroc} (0.1-0.18s vs 0.5-5s). Our secondary backend latter utilizes a classical loop closure mechanism, which employs image retrieval and pose graph optimization to correct for scale drift, and runs on the CPU.

\section{Related Work}

\smallskip\noindent\textbf{Zero-Shot Cross-Domain Generalization} is a longstanding problem in visual SLAM. The challenge is to develop a system which avoids catastrophic failures in different domains, without requiring re-training. Classical approaches to SLAM~\cite{orbslam1,orbslam2,orbslam3} are prone to catastrophic failures during fast camera motion, and generally underperform deep approaches on indoor datasets~\cite{dso, gao2018ldso}. Several works have proposed learning a generalized system for SLAM by using a deep-network backbone trained entirely on synthetic data. TartanVO~\cite{wang2021tartanvo} trained on TartanAir~\cite{wang2020tartanair} and showed strong performance on both indoor and outdoor settings without fine tuning. DROID-SLAM~\cite{teed2021droid} and DPVO~\cite{teed2023deep} followed a similar approach, but used a differentiable bundle adjustment layer in order to learn outlier rejection by supervising on the predicted camera poses.

Our approach is also trained only on synthetic data, however we demonstrate better generalization and/or runtime. Many works in VO/SLAM focus on in-domain accuracy, i.e., approaches trained or developed with a particular test setting taken into special consideration~\cite{dfvo, wang2017deepvo, wang2019improving, yin2018geonet, li2018undeepvo, xu2021attention, yang2020d3vo, shen2023dytanvo, zhang2020vdo}. This setting is orthogonal to ours; we do not claim to outperform VO/SLAM systems specialized for autonomous driving applications on the KITTI dataset, for example.\looseness=-1

\smallskip\noindent\textbf{Monocular SLAM} is especially challenging due to the ambiguity of scale in monocular video. Several VO/SLAM works remove the scale ambiguity altogether by relying on stereo video, inertial measurements, or depth~\cite{fu2023islam, schops2019bad, schneider2018maplab}. In contrast, our method operates on monocular video. We focus our evaluation on methods which do the same. Monocular SLAM is important due to the wide availability of monocular video, and because they can be easily adapted to use additional sensors, whereas the reverse is not always true. 

\smallskip\noindent\textbf{Neural SLAM and rendering-focused SLAM}: Several recent approaches use Gaussian-splatting and/or NeRFs~\cite{sucar2021imap, zhu2022nice, keetha2023splatam, li2023dense}. These rendering-based approaches are primarily designed for high-quality reconstruction/rendering, with tracking being a secondary focus usually only evaluated with smooth/slow camera motion on \cite{straub2019replica} or \cite{dai2017scannet}. In contrast, we focus on tracking accuracy in hard settings, similar to \cite{teed2021droid,teed2023deep,orbslam1,orbslam2,orbslam3,dso}. \cite{goslam} is an exception, which we compare to.\looseness=-1%

\smallskip\noindent\textbf{Loop Closure} enables VO/SLAM methods to correct drift by adding factors between temporally-distant pose variables. Campos et al.~\cite{orbslam3} categorized the types of loop closure as \emph{mid-term} and \emph{long-term} data-association based on their approach to detecting loops and optimizing the scene graph. \emph{Mid-term} uses the current estimate of poses and depth to detect loops, and updates them using bundle adjustment. \emph{Long-term} uses visual place recognition to detect loops and updates the state using pose graph optimization. DROID-SLAM~\cite{teed2021droid} uses mid-term. LDSO~\cite{gao2018ldso} uses long-term. VO systems use neither (by definition). ORB-SLAM~\cite{orbslam3} uses both. DPV-SLAM uses mid-term, or optionally both.

\smallskip\noindent\textbf{Deep Patch Visual Odometry (DPVO)} was proposed as a faster alternative to the visual odometry from DROID, based on two insights. The first is that virtually every approach to SLAM provides some mechanism to trade-off accuracy for speed and/or memory (to an extent). For example, one can increase the number of keypoints, RANSAC iterations, the optimization window size, the image resolution, the number of keyframes produced, or the connectivity of the factor graph. For SLAM methods with deep-network backbones, one can also increase the feature dimension, add more layers, or use quantization / mixed-precision.

The second insight is that, by predicting sparse optical flow as opposed to dense, the resulting memory/runtime savings are sufficient to offset the initial accuracy loss by \emph{spending} them in other aspects of the design. This allows DPVO to achieve similar accuracy to DROID-SLAM's frontend, with much lower cost and faster inference. The drawback is that the DPVO design is more challenging to adapt to a full SLAM system due to the large per-frame storage requirement. DPVO also suffers from the same performance issues as DROID-SLAM on outdoor datasets.%

\section{Approach}

\subsection{DPVO Preliminaries}\label{sec:preliminaries}

Our system is based on Deep Patch Visual Odometry (DPVO)~\cite{teed2023deep}. DPVO is a sparse analog of the visual odometry frontend of DROID-SLAM which achieves similar accuracy with much lower latency and memory. In this section, we will discuss the details of DPVO that are relevant to our contribution. For more details, we refer the readers to the original DPVO paper.%

\smallskip\noindent\textbf{DPVO Overview:} Given an input video stream, DPVO seeks to estimate the 2D motion of selected keypoints across time by predicting optical flow and updating the depth and camera poses using bundle adjustment. DPVO only supports visual odometry, so it operates on a sliding window of frames and removes keyframes and features once they fall outside of the optimization window.%

\begin{figure*}[th]
    \centering
    \includegraphics[width=1.0\textwidth]{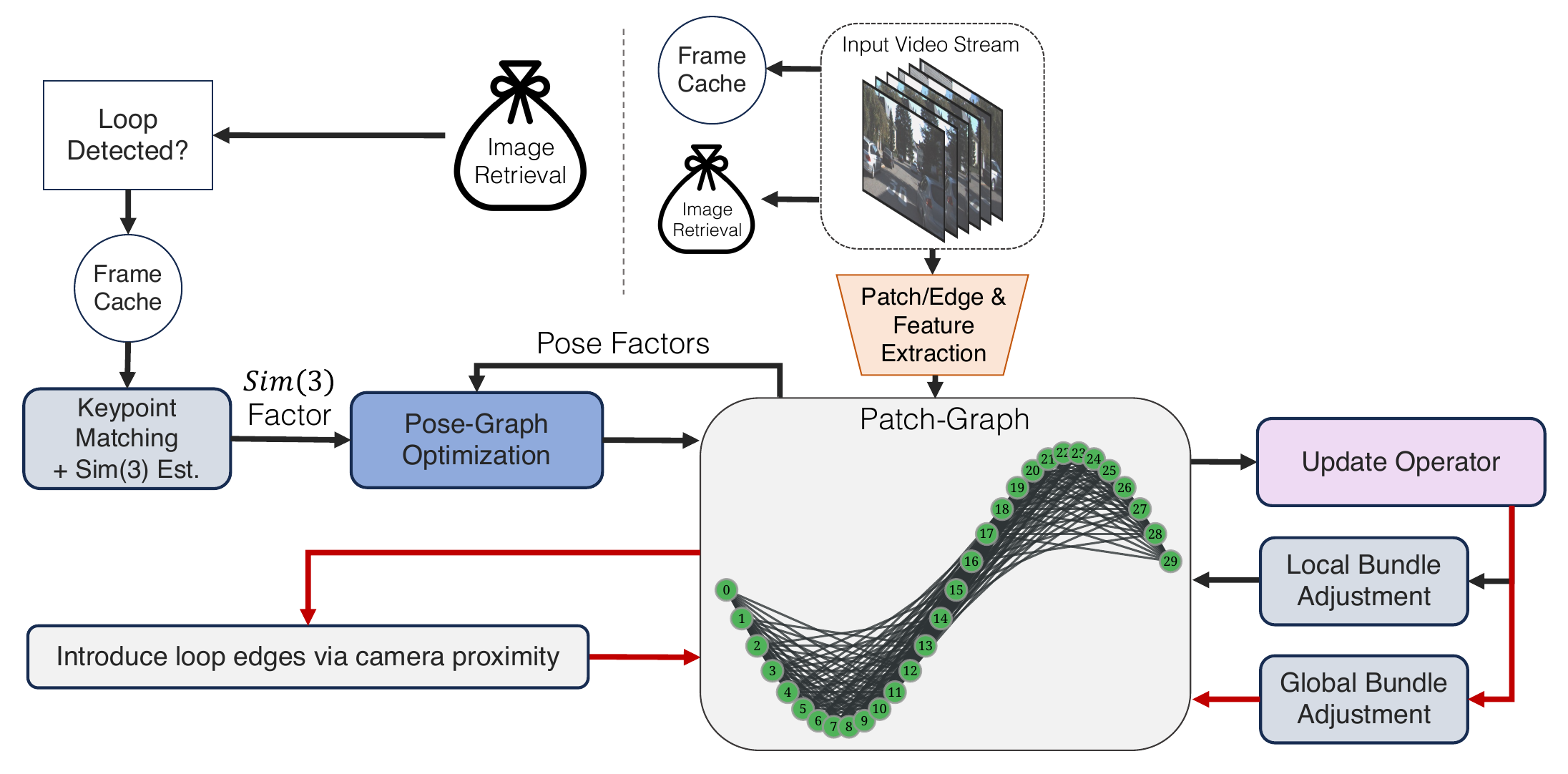}
    \caption{Overview of DPV-SLAM. Our system is based on the odometry system from DPVO~\cite{teed2023deep}, and introduces two efficient loop closure mechanisms to correct for accumulated drift. Like DPVO, our system utilizes a patch graph scene representation, and alternates between predicting sparse optical flow residuals and optimizing the camera poses and depth using bundle adjustment. Our proximity loop closure detects loops using the pre-estimated geometry and jointly refines all variables together. Our classical loop-closure uses image retrieval and pose graph optimization.}
    \label{fig:system_overview}
\end{figure*}

\begin{figure*}[t]
    \centering
    \includegraphics[width=1.0\textwidth]{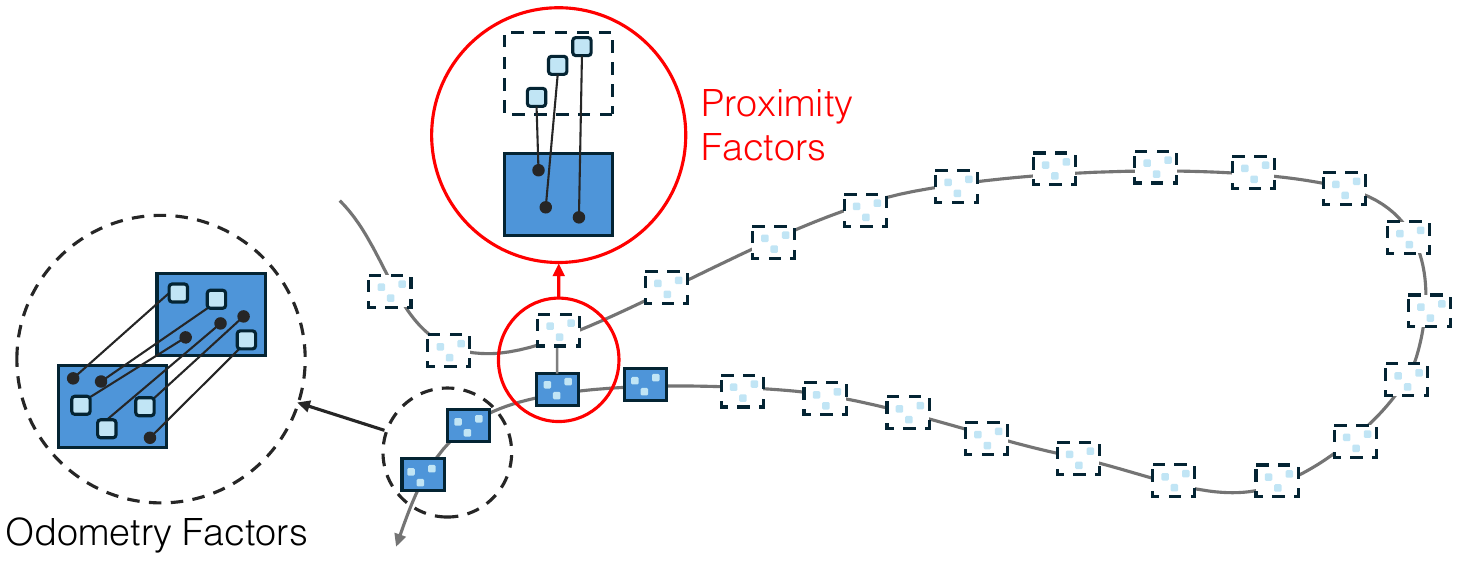}
    \caption{The patch graph for DPV-SLAM. We introduce directed edges from old frames to recent frames still in-use by the odometry component. These edges are chosen based on the camera's \emph{proximity} to previously visited locations. The construction of this patch graph only requires storing a finite number of dense feature maps, keeping the overall memory consumption small.}
    \label{fig:midterm_patchgraph}
\end{figure*}

\smallskip\noindent\textbf{The patch graph:} DPVO uses a scene representation known as a patch graph, in which each frame $i$ contains a set of $p \times p$ patches $\mathbf{P}_{ik}$
\begin{equation}
    \mathbf{P}_{ik} = \left( \begin{array}{c}
    \mathbf{x} \\ \mathbf{y} \\ \mathbf{1} \\ \mathbf{d}
\end{array} \right) \qquad \mathbf{x},\mathbf{y},\mathbf{d} \in \mathbb{R}^{1 \times p^2}
\end{equation}
where $\mathbf{d}$ is the inverse depth estimate. We denote the number of patches for frame $i$ as $K_i$, and $[K_i] := \{1,...,K_i\}$. The patch graph is a bipartite graph, in which directed edges connect patches to frames. The scene representation is used by DPVO and DPV-SLAM. We visualize an example of our patch graph in Fig.~\ref{fig:midterm_patchgraph}.

Given the current poses, inverse depths and camera intrinsics, we can reproject any patch to any other frame. We denote the set of edges as $F$, the global camera pose for frame $i$ as $G_i$, and the 3D$\rightarrow$2D pinhole-projection function as $\Pi(\cdot)$. We represent edges from $\mathbf{P}_{ik}$ to frame $j$ as $(i,k,j)$. The reprojection of $\mathbf{P}_{ik}$ to frame $j$ is denoted as
\begin{equation}
    \label{eq:reprojection}
    \mathbf{P}^{\prime}_{ikj} = \Pi[G_{j}^{-1} \cdot G_i \cdot \Pi^{-1}(\mathbf{P}_{ik})]
\end{equation}
The explicit objective of DPVO is to predict residual updates $\Delta_{ikj}$ to $\mathbf{P}^{\prime}_{ikj}$ for all edges in order to improve the visual alignment. The resulting $\mathcal{I}_{ikj}~:=~(\mathbf{P}^{\prime}_{ikj}~+~\Delta_{ikj})$ represents the model's \emph{ideal} reprojection of $\mathbf{P}_{ik}$ into frame $j$. After predicting $\Delta_{ikj}$, DPVO optimizes the patch depths and camera poses to align the actual patch reprojections to the ideal reprojections.
\begin{equation}
\label{eq:bundleadjustment}
\argmin_{G, \mathbf{d}} \sum_i \sum_{k \in [K_i]} \sum_{j: (i,k,j) \in F}  \norm{\Pi[G_{j}^{-1} \cdot G_i \cdot \Pi^{-1}(\mathbf{P}_{ik})] - \mathcal{I}_{ikj}}^2_{\Sigma_{ikj}}
\end{equation}
Note that in eq.~\ref{eq:bundleadjustment}, $\mathcal{I}_{ikj}$ is treated as a constant. In addition to $\Delta_{ij}$, DPVO also predicts a confidence estimate $w_{ikj} \in \mathbb{R}^2$ for each edge. Eq.~\ref{eq:bundleadjustment} minimizes the Mahalanobis distance, in which the error terms are weighted by the predicted confidences: $\Sigma_{ikj} = diag(w_{ikj})$.

\smallskip\noindent\textbf{Patch extraction:} DPVO selects keypoints randomly, as opposed to the usual strategy of using a detector or something more sophisticated~\cite{dso}. This counter-intuitive strategy works surprisingly well~\cite{teed2023deep}, and is trivial to implement. The patch features are cropped around the chosen 2D keypoints from dense $(\nicefrac{H}{4}\times \nicefrac{W}{4})$ feature maps predicted by a residual network learned end-to-end with the full model. DPVO extracts both $1\times1$ context features, and $p\times p$ correlation features. The correlation features are used to evaluate the visual alignment of the current pose and depth estimates, whereas the context features are provided as-is to the update operator.%

\smallskip\noindent\textbf{Update operator:} The update operator of DPVO is the recurrent module used to predict $\Delta_{ikj}$ and $w_{ikj}$ for all edges in the patch graph. As an RNN, it also maintains a running hidden state $h_{ikj} \in \mathbb{R}^{384}$ for every edge $(i,k,j)$. The architecture includes several fully-connected gated residual units. As input, the update operator accepts the previous hidden state $h_{ikj}$, correlation features $\mathbf{C}_{ikj}$, and the context features for $\mathbf{P}_{ik}$.

\begin{figure*}[t]
    \centering
    \includegraphics[width=1.0\textwidth]{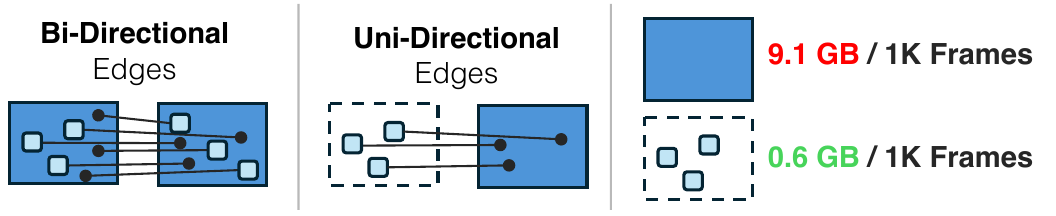}
    \caption{Each directed edge in the patch graph introduces a highly-asymmetric memory overhead for the source patch and destination frame, respectively. Each edge is also associated with its own reprojection-error factor in the optimization which constrains both camera poses. This means we can flip their direction arbitrarily to influence which frames incur the memory cost, without significantly impacting the optimization result. This is a unique property of DPVO's patch representation and motivates the use of uni-directional edges; prior works used bi-directional (and dense) edges in their backends.}
    \label{fig:directional_cost}
\end{figure*}

\smallskip\noindent\textbf{Patch correlation:} Correlation features $\mathbf{C}_{ikj}$ are computed for each edge in order to evaluate the visual alignment produced by the current depth and pose estimates. To compute $\mathbf{C}$, we use eq.~\ref{eq:reprojection} to reproject $\mathbf{P}_{ik}$ into frame $j$. Let $\mathbf{g}(u,v)$ represent the $p\times p$ patch correlation features indexed at $(u,v)$ and $\mathbf{P}^{\prime}(u,v)$ denote $\mathbf{P}_{ijk}^{\prime}$ indexed at $(u,v)$. $f(\cdot)$ denotes bilinear sampling, $\langle\cdot\rangle$ a dot product, and $\Delta_{\alpha\beta}$ a $7\times7$ grid centered at 0, indexed at $(\alpha, \beta)$. Each value in $\mathbf{C}\in \mathbb{R}^{p\times p \times 7 \times 7}$ is computed as:
\begin{equation}
    \label{eq:correlation}
    \mathbf{C}(u,v,\alpha,\beta) = \langle \mathbf{g}(u,v), \ \mathbf{f}(\mathbf{P}'(u,v) + \Delta_{\alpha\beta})
    \rangle
\end{equation}
$\mathbf{C}$ is recomputed and flattened before each invocation of the update operator. Colloquially, eq.~\ref{eq:correlation} computes a dot product between correlation features for all pairs of grid points around either end of the flow vector induced from the poses and $\textbf{P}_{k}$. The key insight is that eq.~\ref{eq:correlation} requires storing the full correlation feature map for frame $j$ in memory, since $\textbf{P}^{\prime}_{ikj}$ is unbounded. This is an expensive requirement, which is addressed in our proximity loop-closure.

\subsection{Proximity Loop Closure}
\label{sec:proximityloopclosure}

We contribute a loop closure mechanism to DPVO which detects loops via the cameras' proximity to previously visited locations. This seeks to improve global consistency by periodically inserting long-range edges into the patch graph, updating their optical flow, and performing global bundle adjustment to update all camera poses and depth. Previous deep SLAM systems~\cite{goslam, teed2021droid} require two GPUs in order to consistently average real-time inference. This is because CUDA operations typically utilize all available cores on their host device, and therefore must run sequentially, even when they are being called from separate processes. All Pytorch/CUDA operations in DPV-SLAM run in a single process on a single device, which is both simpler and computationally cheaper.

\smallskip\noindent\textbf{Constructing the Patch Graph:} DPVO, like DROID-SLAM~\cite{teed2021droid}, stores dense feature maps in memory for updating the optical flow predictions. The larger the optimization window, the more feature maps must be kept, increasing cost. This problem is exacerbated in DPVO, where the feature maps are twice the spatial resolution ($\nicefrac{1}{4}$ vs $\nicefrac{1}{8}$ of the frame size). %

We observe that, for each directed edge in the patch graph, the correlation operator only requires storing dense features for the destination frame. Additionally, DPV-SLAM borrows the inverse depth parametrization~\cite{civera2008inverse} for bundle adjustment, so each factor in the optimization will constrain both the source and destination camera poses regardless of its direction (see eq.~\ref{eq:bundleadjustment}). This means we can arbitrarily flip any edge in the patch graph to influence which dense features must be stored, without significantly affecting the optimization result.

We exploit this fact to minimize the number of deep features stored in memory. Specifically, we permanently store only the patch features for all previous time-steps and create uni-directional edges connecting these patches to recently observed frames. Consequently, we incur only a minor storage overhead from the patch features ($\approx 0.6$G / 1K frames). We depict this globally-connected patch graph in Fig.~\ref{fig:midterm_patchgraph}.%

\smallskip\noindent\textbf{Efficient Global Optimization:} We mix both odometry and loop-closure factors in the same optimization. This mandates a bundle adjustment implementation which can handle global optimization without severely impacting the odometry component. DPVO's pre-existing bundle adjustment is GPU-accelerated, but is still inefficient for large, sparse optimization problems. This is precisely the challenge we face when introducing a small number of long-range proximity factors into the optimization. A viable solution is to leverage the sparse structure of the hessian by implementing bundle adjustment in CUDA with block sparse representations. While prior work has implemented such a system for dense, uniformly sized depth maps~\cite{teed2021droid}, such a system has not been built for sparse, varying sized patch graphs. We contribute our implementation and use it to perform efficient global optimization.

\begin{figure*}[t]
    \centering
    \includegraphics[width=1.0\textwidth]{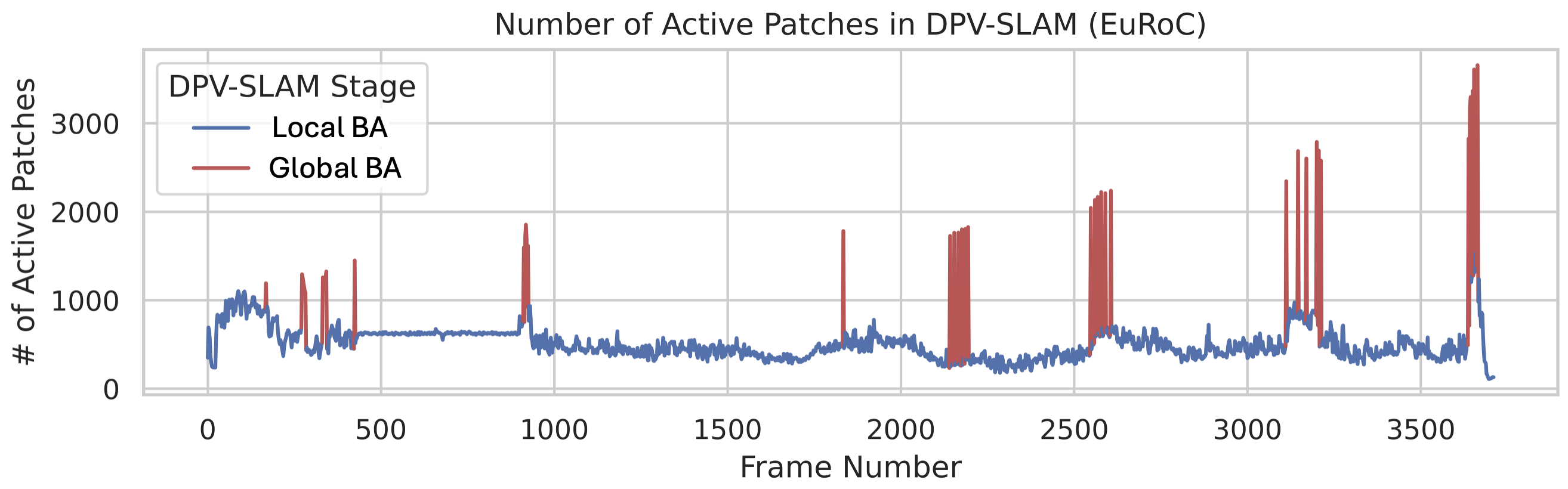}
    \caption{We visualize the number of patches participating in the optimization over the coarse of a video. During invocations of our proximity loop-closure, we perform global bundle adjustment which updates a significant portion of patch depths. Here, we only consider patches with at least one high-confidence outgoing edge $(w > 0.5)$.}
    \label{fig:todo}
\end{figure*}

\subsection{Classical Loop Closure}
\label{sec:classicalloopclosure}
\begin{figure*}[t]
    \centering
    \includegraphics[width=0.7\textwidth]{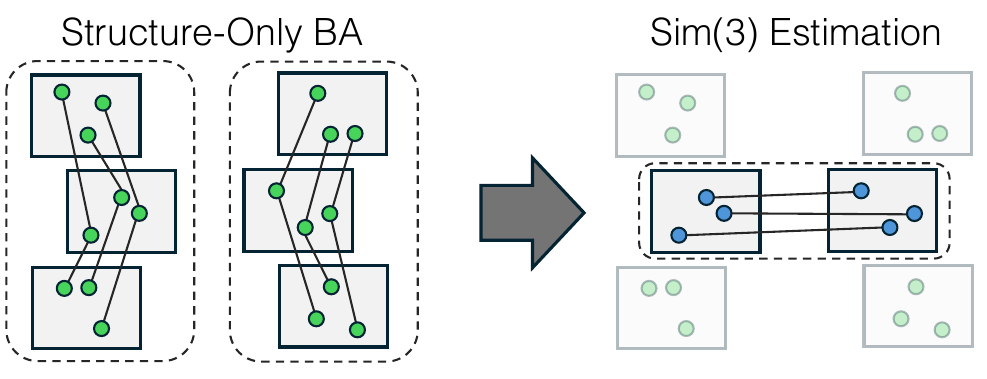}
    \caption{Drift estimation. After identifying a candidate image pair for loop closure using image retrieval, we seek to estimate the accumulated drift as a relative 7DOF transformation. Using off-the-shelf detectors and matchers, we estimate 2D correspondence from each retrieved image to its two temporal neighbors and perform structure-only bundle adjustment to triangulate their depth. Finally, we match between the resulting 3D keypoints and estimate a 7DOF point-cloud alignment with RANSAC+Umeyama~\cite{umeyama1991least}.}
    \label{fig:sim3_est}
\end{figure*}

Separate from our proximity loop closure, we also support a more traditional SLAM backend which uses classical image retrieval and pose graph optimization. This type of loop closure is better equipped to recover from scale drift. We refer to the variant of our model with both proximity and classical loop closure as DPV-SLAM++.\looseness=-1

\smallskip\noindent\textbf{Detecting Loops:} We identify candidate image pairs using dBoW2~\cite{dbow2} for image-retrieval, which requires extracting ORB~\cite{orb} features for each frame. The process of extracting features, indexing and searching the DBoW model takes less time than the forward pass of DPVO and happens concurrently in a separate process, thereby incurring virtually 0 runtime overhead. Following prior work~\cite{orbslam3}, we look for multiple consecutive defections to increase precision.

\begin{figure*}[ht]
    \centering
    \includegraphics[width=\textwidth]{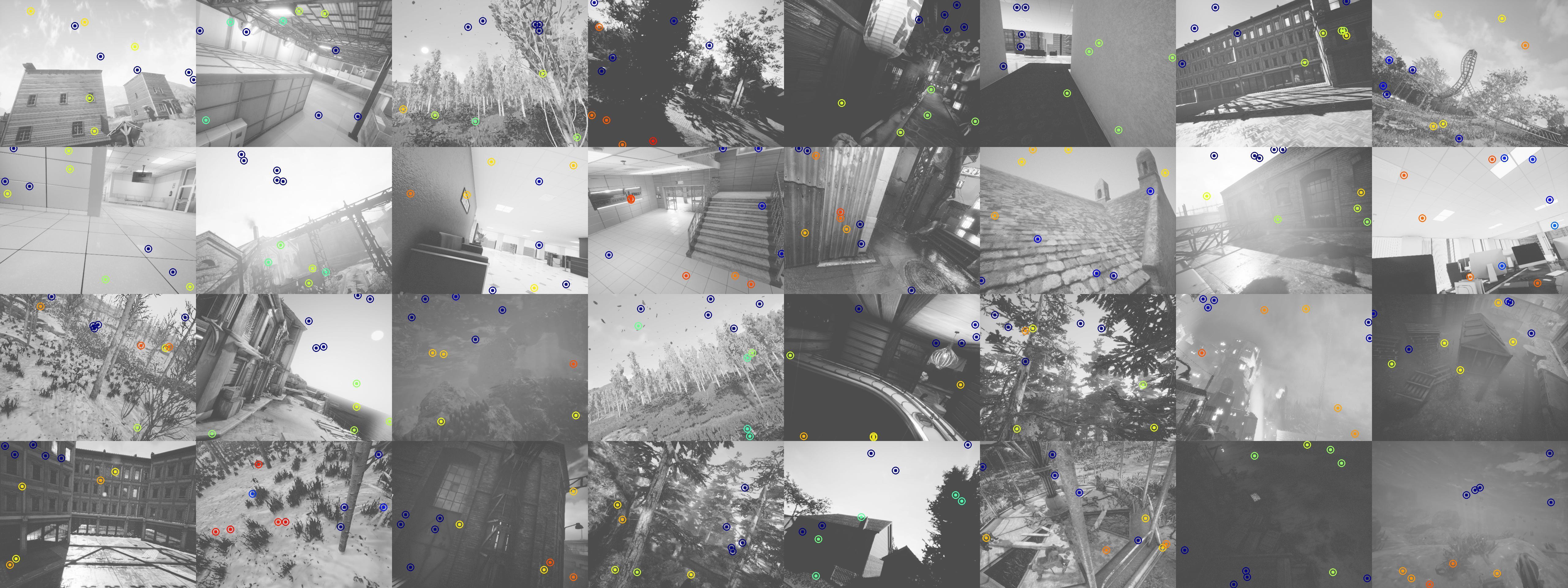}
    \caption{Visualization of the 5 most confident (red) and 5 least confident (blue) patch centroids on TartanAir~\cite{wang2020tartanair}. 96 keypoints are chosen randomly in each image. Since only the edges have associated weights, we compute the ``confidence'' of each keypoint as the maximum predicted weight over all of its outgoing edges. In contrast to the usual expectation that the most salient features are the easiest to track, we observe that DPV-SLAM often prefers points on near-featureless regions.} %
    \label{fig:detection}
\end{figure*}

\smallskip\noindent\textbf{Estimating Drift:} We week to estimate the accumulated drift $\Delta S^{loop}_{jk} \in Sim(3)$ between each retrieved image pair $(j,k)$. This is often done by matching between their previously-mapped keypoints. However, our keypoints cannot directly be used for matching since DPVO does not use a repeatable-keypoint detector. This decision has been justified in several prior works ~\cite{teed2023deep, gao2018ldso, dso} which showed that such detectors are not ideal for tracking small-motions. See Fig. ~\ref{fig:detection} for examples.

We instead leverage off-the-shelf keypoint detectors and matchers~\cite{riba2020kornia, tyszkiewicz2020disk, lindenberger2023lightglue}, only during loop closure, to estimate 2D correspondence from each retrieved image to its temporal neighbors and perform structure-only bundle adjustment to triangulate their depth. We then match the 3D keypoints between the chosen image pair and align their two point clouds using RANSAC+Umeyama~\cite{umeyama1991least}. We depict this process in Fig.~\ref{fig:sim3_est}. \smallskip

\begin{figure}[ht]
\centering
\begin{subfigure}[b]{0.4\textwidth}
    \centering
    \includegraphics[width=\textwidth]{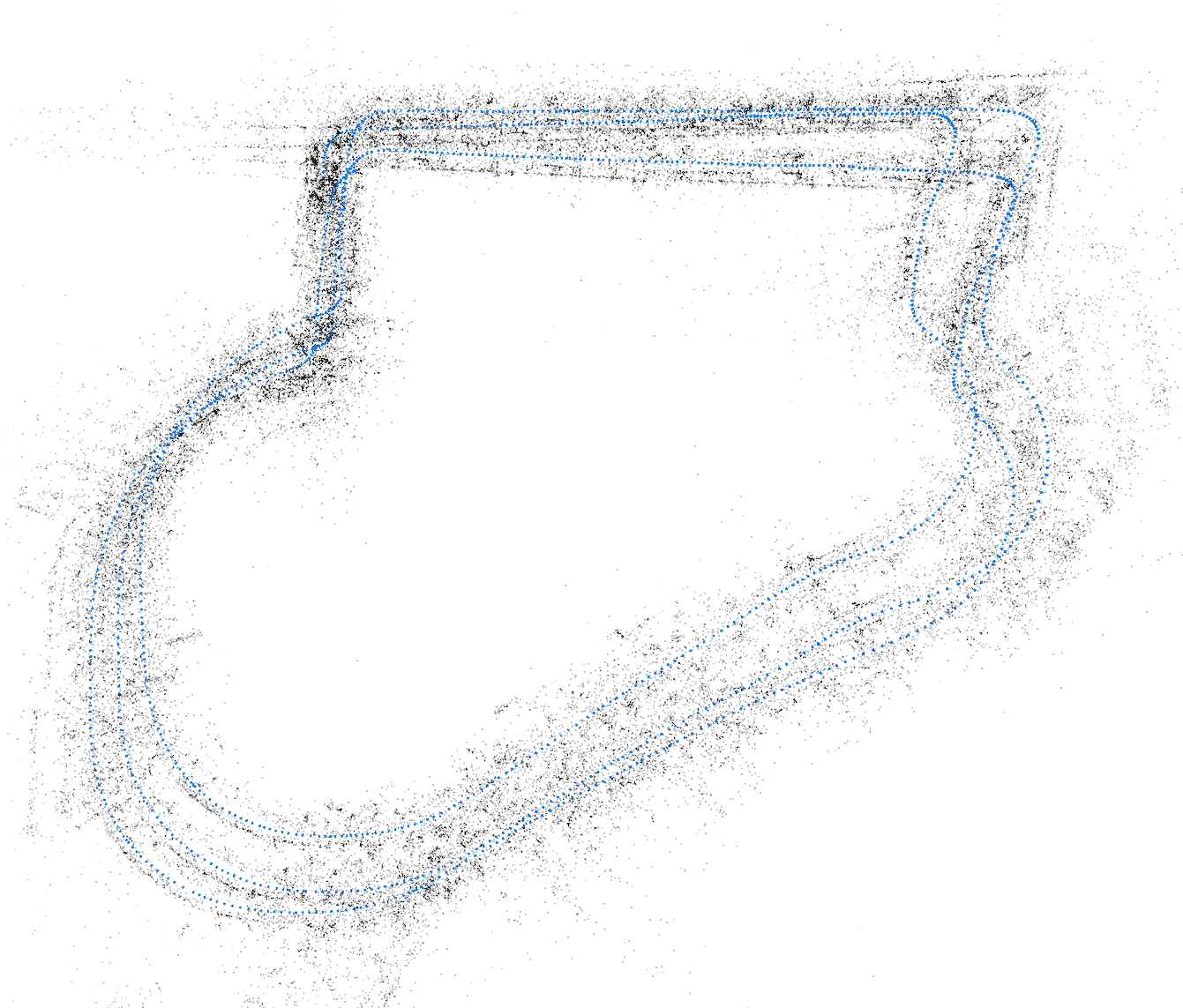} 
    \caption{DPVO}
    \label{fig:4seasonsnoloop}
\end{subfigure}
\begin{subfigure}[b]{0.4\textwidth}
    \centering
    \includegraphics[width=\textwidth]{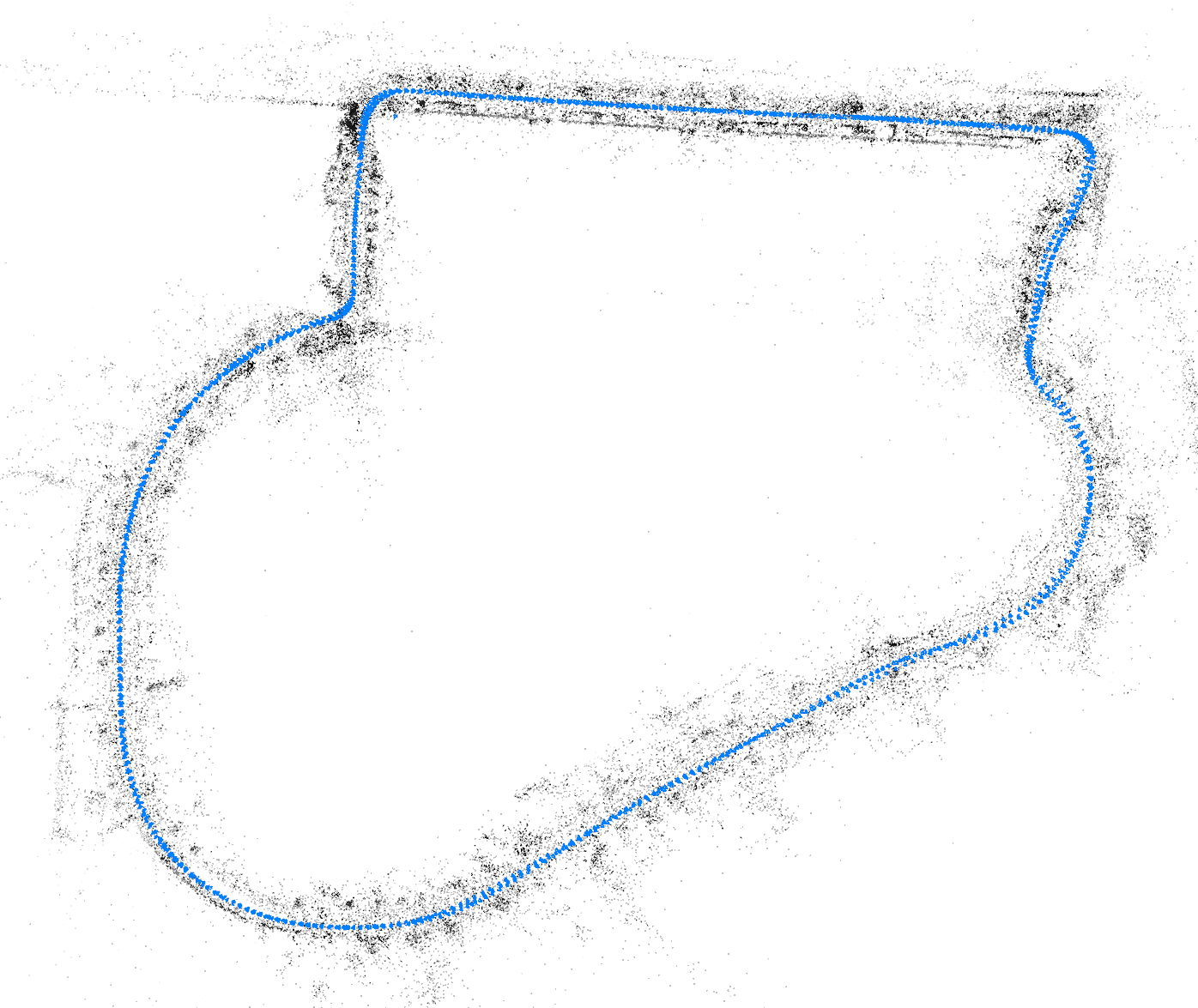} 
    \caption{DPVO with classical loop closure}
    \label{fig:4seasonsyesloop}
\end{subfigure}
\caption{Predictions of the DPVO~\cite{teed2023deep} base model with and without our classical loop closure (Sec.~\ref{sec:classicalloopclosure}) on the ``Business Campus" sequence of the 4Seasons dataset~\cite{wenzel20214seasons}.}
\end{figure}

\noindent\textbf{Optimization:} The final step is to estimate an absolute similarity for all keyframes using a simplified version of the algorithm from \cite{strasdat2010scale}. We assign each keyframe $i$ an initial absolute similarity $S_i\in Sim(3)$ by converting their current global pose estimates into similarities with scale 1. These are the free variables of the optimization, while the relative-similarity terms denoted with the $\Delta$ suffix are constants. We add a smoothness term to our optimization objective between each keyframe and its successor, defined in the tangent space of $Sim(3)$:
\begin{equation}
    r_i = log_{Sim(3)}\big(\Delta S_{(i,i+1)}^{-1} \cdot S_i^{-1} \cdot S_{i+1}\big)
\end{equation}
and error terms for the estimated loop connections:
\begin{equation}
    r_{jk} = log_{Sim(3)} \big(\Delta S^{loop}_{jk} \cdot S^{-1}_{j}\cdot S_{k}\big)
\end{equation}
We then optimize the following objective using the Levengberg-Marquardt algorithm:
\begin{equation}
    \label{eq:pgo}
    \argmin_{S_1,...S_N} \Bigg( \sum_i^N \norm{r_{i}}_2^2 + \sum^L_{(j,k)} \norm{r_{jk}}_2^2 \Bigg)
\end{equation}
where $N$ is the total number of keyframes and $L$ is the list of detected loops. For each absolute similarity $S_i = (t_i, R_i, s_i)$, the global poses and inverse depths are updated as $G_i \leftarrow (t_i, R_i)$ and $ d_i \leftarrow d_i/s_i$. The pose graph optimization is performed on the CPU in parallel to the main process, thereby incurring negligible runtime overhead.
\begin{figure}[htbp]
\centering
\begin{subfigure}[b]{0.33\textwidth}
    \centering
    \includegraphics[width=\textwidth]{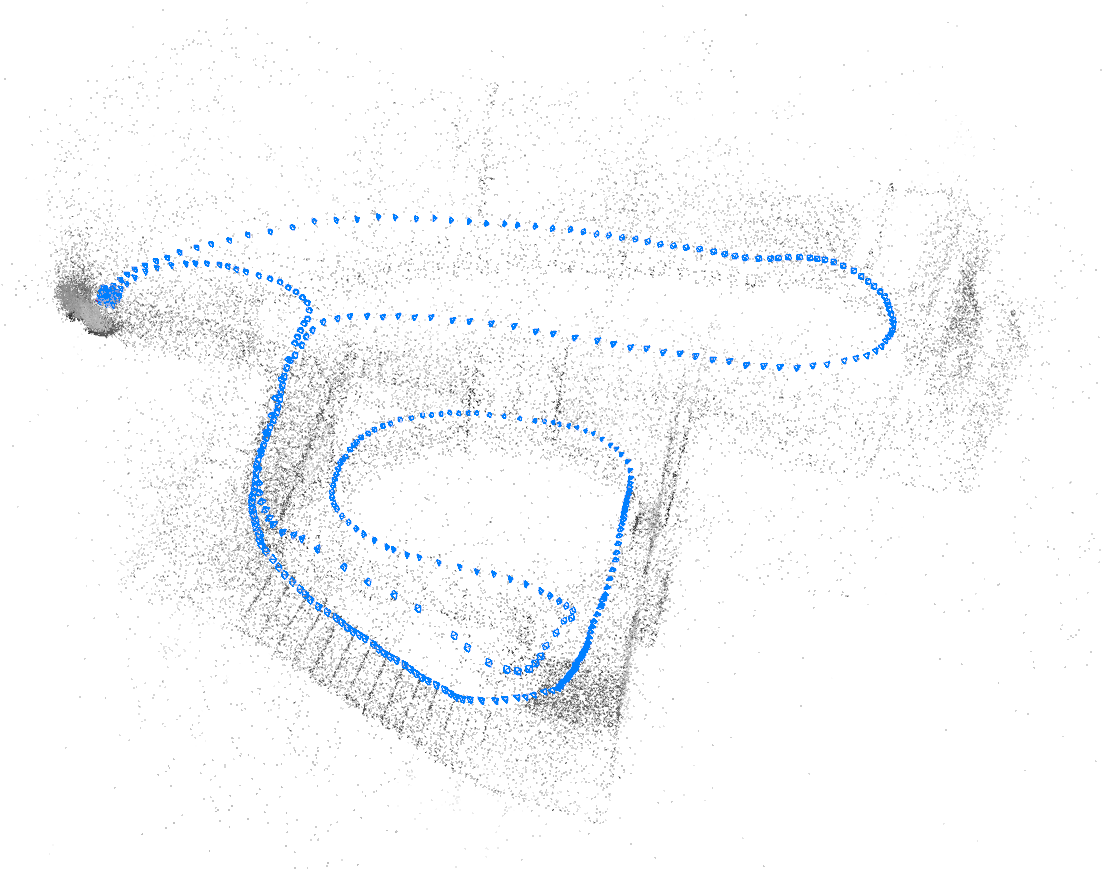} 
    \caption{Sequence 41}
    \label{fig:todo}
\end{subfigure}
\begin{subfigure}[b]{0.32\textwidth}
    \centering
    \includegraphics[width=\textwidth]{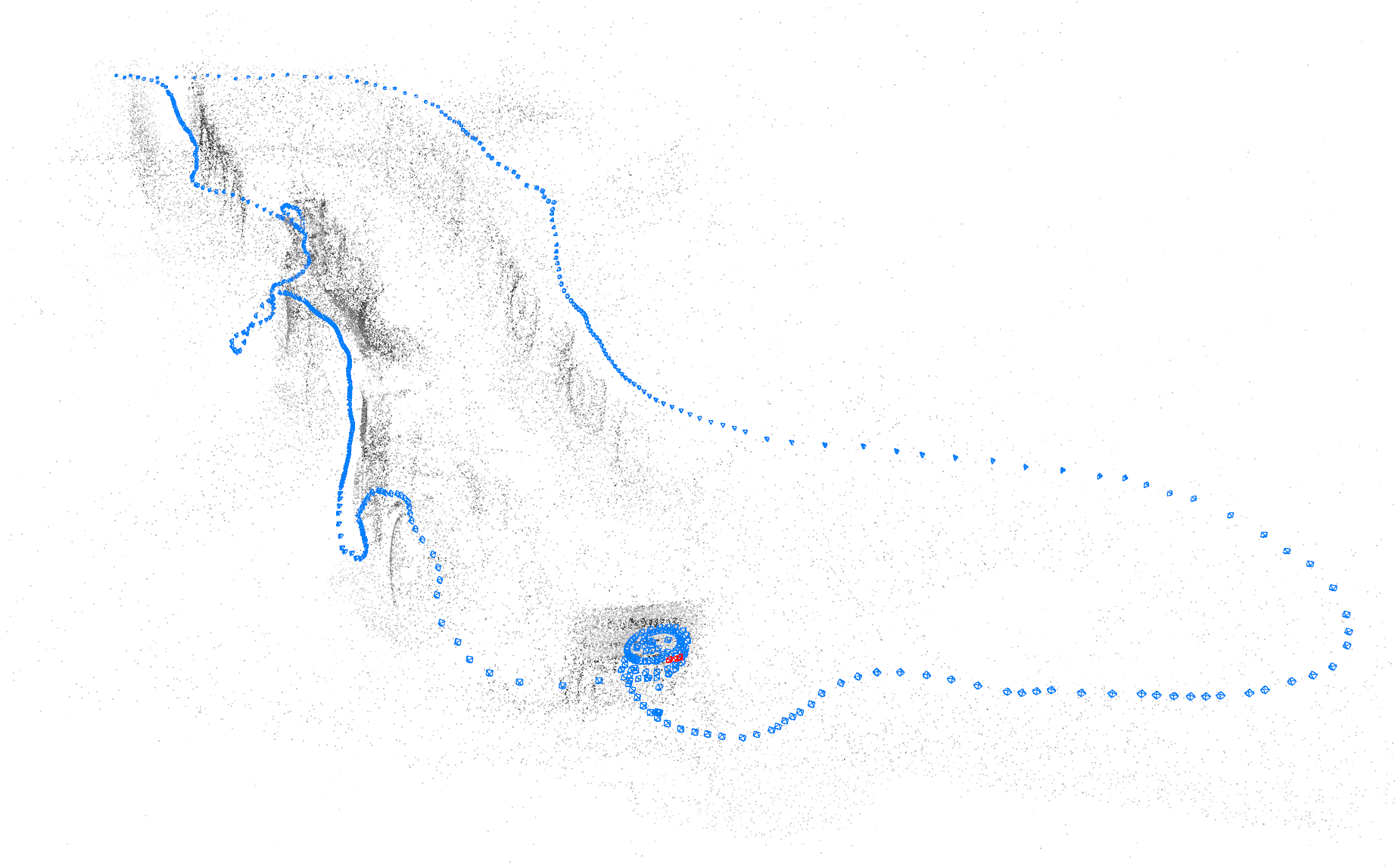} 
    \caption{Sequence 20}
    \label{fig:todo}
\end{subfigure}
\begin{subfigure}[b]{0.33\textwidth}
    \centering
    \includegraphics[width=\textwidth]{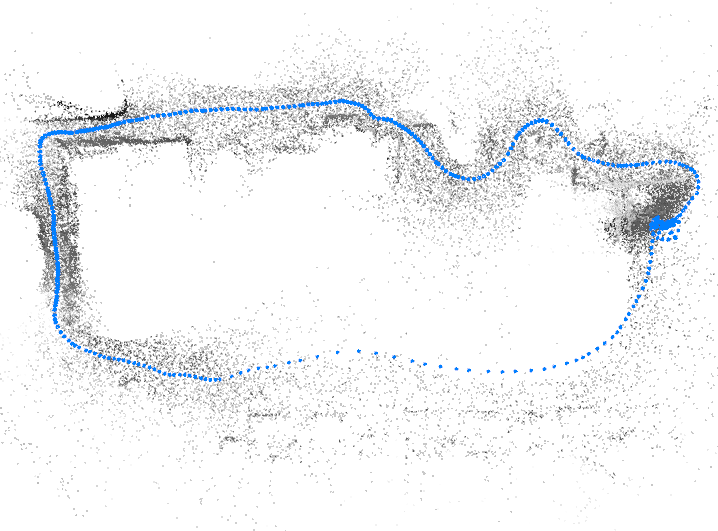} 
    \caption{Sequence 43}
    \label{fig:todo}
\end{subfigure}
\caption{Qualitative visualization on TUM-Mono~\cite{engel2016photometrically}}
\end{figure}

\section{Experiments}

We evaluate DPV-SLAM on four datasets: KITTI~\cite{geiger2012we}, TUM-RGBD~\cite{tumrgbd}, EuRoC-MAV~\cite{burri2016euroc}, and the TartanAir~\cite{wang2020tartanair} test set from the ECCV 2020 SLAM competition. We compare DPV-SLAM to methods which report results on both indoor and outdoor settings, and that do not require re-training their model per-domain. DPV-SLAM experiments were run 5 times, and we report the median result. $\mathsection$ denotes values which we measured using their open source code, since they were not reported in the original paper. For GO-SLAM, timings were measured using tracking/mono mode. All timing experiments were performed on an RTX-3090. The ``DPV-SLAM'' and ``DPV-SLAM++'' experiments both utilize the proximity loop closure mechanism. The latter also uses the classical loop closure.

\smallskip\noindent\textbf{TUM-RGBD~\cite{tumrgbd}} We evaluate monocular SLAM on the entirety of the Freiburg 1 set of the TUM-RGBD benchmark in Tab.~\ref{table:TUM}. This benchmark evaluates indoor, handheld camera motion, and is especially challenging due to rolling shutter effects and motion blur. Video is recorded at 30FPS.

\begin{table}[t]
\centering
\resizebox{\linewidth}{!}{%
\begin{tabular}{l|ccccccccc | c | c | c}
& 360 & desk & desk2 & floor & plant & room & rpy & teddy & xyz & Avg & FPS & VRAM \\
\toprule
ORB-SLAM2~\cite{orbslam2} & X & 0.071 & X & 0.023 & X & X & X & X & 0.010 & - \\
ORB-SLAM3~\cite{orbslam3} & X & {0.017} & 0.210 & X & 0.034 & X & X & X & {\textbf{0.009}} & - \\
DeepTAM~\cite{zhou2018deeptam} & {\textbf{0.111}} & 0.053 & 0.103 & 0.206 & 0.064 & 0.239 & 0.093 & 0.144 & 0.036 & 0.116\\
TartanVO~\cite{wang2021tartanvo} & 0.178 & 0.125 & 0.122 & 0.349 & 0.297 & 0.333 & 0.049 & 0.339 & 0.062 & 0.206\\
DeepV2D~\cite{teed2018deepv2d} & 0.243 & 0.166 & 0.379 & 1.653 & 0.203 & 0.246 & 0.105 & 0.316 & 0.064 & 0.375 \\
DeepV2D [TartanAir] & 0.182 & 0.652 & 0.633 & 0.579 & 0.582 & 0.776 & 0.053 & 0.602 & 0.150 & 0.468 \\
DeepFactors~\cite{czarnowski2020deepfactors} & 0.159 & 0.170 & 0.253 & 0.169 & 0.305 & 0.364 & 0.043 & 0.601 & {0.035} & 0.233\\
DeFlowSLAM~\cite{deflowslam} & 0.159 & \textbf{0.016} & 0.030 & 0.169 & 0.048 & 0.538 & \textbf{0.021} & \textbf{0.039} & \textbf{0.009} & 0.114 \\
GO-SLAM~\cite{goslam} & 0.089 & \textbf{0.016} & \textbf{0.028} & 0.025 & 0.026 & 0.052 & 0.019 & 0.048 & 0.010 & \textbf{0.035} & $6.4^\mathsection$ & 7.2G$^\mathsection$\\
DROID-SLAM~\cite{teed2021droid} & \textbf{0.111} & 0.018 & 0.042 & {0.021} & \textbf{0.016} & \textbf{0.049} & {0.026} & 0.048 & {0.012} & {0.038} & \textbf{30} & 8.5G$^\mathsection$\\
\midrule
DPV-SLAM & 0.112 & 0.018 & {0.029} & {0.057} & {0.021} & {0.330} & {{0.030}} & 0.084 & 0.010 & {0.076} & \textbf{30} & \textbf{4.0G}\\
DPV-SLAM++ & 0.132 & 0.018 & 0.029 & 0.050 & 0.022 & 0.096 & 0.032 & 0.098 & 0.010 & 0.054 & \textbf{30} & {6.0G}\\
\bottomrule
\end{tabular}
}
\caption{ATE on the TUM-RGBD benchmark. Bolded indicates the best result.  We report runtimes and memory for methods whose average error is similar to ours. We perform similarly to other methods based on DROID-SLAM~\cite{deflowslam, goslam, teed2021droid} (0.054-0.076 vs 0.035-0.114), however these approaches are more expensive (4.0-6.0G vs 7.2-8.5G). The results in this table, combined with Tab~\ref{table:kitti}, indicate that our approach is robust both indoors and outdoors.}

\label{table:TUM}
\end{table}

\begin{table}[h!]
\begin{subtable}{\linewidth}\centering
\resizebox{\linewidth}{!}{%
\setlength{\tabcolsep}{7pt}
\begin{tabular}{l| cccccccccc}
\toprule
Sequence & \multicolumn{2}{c}{06} & \multicolumn{2}{c}{07} & \multicolumn{2}{c}{09} & \multicolumn{2}{c}{10} & \multicolumn{2}{c}{Avg} \\
& $t_{rel}$ & $r_{rel}$ & $t_{rel}$ & $r_{rel}$ & $t_{rel}$ & $r_{rel}$ & $t_{rel}$ & $r_{rel}$ & $t_{rel}$ & $r_{rel}$ \\
\toprule
TartanVO~\cite{wang2021tartanvo}  & \textbf{4.72} & 2.95 & 4.32 & 3.41 & \textbf{6.0} & 3.11 & 6.89 & 2.73 & \textbf{5.48} & 3.05 \\
DPV-SLAM++ & {4.95} & \textbf{0.16} & \textbf{1.29} & \textbf{0.24} & {17.69} & \textbf{0.23} & \textbf{6.32} & \textbf{0.23} & {7.56} & \textbf{0.22}\\
\bottomrule
\end{tabular}
}
\caption{Comparison with TartanVO\vspace{3mm}}\label{tab:1a}
\end{subtable}%
\\
\begin{subtable}{.99\linewidth}\centering
\resizebox{\linewidth}{!}{%

\begin{tabular}{l| ccccccccccc | c | c}
\toprule
& 00 & 01 & 02 & 03 & 04 & 05 & 06 & 07 & 08 & 09 & 10 & Avg & FPS \\
\toprule
ORB-SLAM2~\cite{orbslam2} & 8.27 & X & \textbf{26.86} & \textbf{1.21} & {0.77} & 7.91 & 12.54 & 3.44 & \textbf{46.81} & 76.54 & \textbf{6.61} & - & 34 \\
ORB-SLAM3~\cite{orbslam3} & \textbf{6.77} & X & 30.500 & 1.036 & 0.930 & 5.542 & 16.605 & 9.700 & 60.687 & 7.899 & 8.650 & - & 34\\
LDSO~\cite{gao2018ldso} & 9.32 & \textbf{11.68} & 31.98 & 2.85 & 1.22 & \textbf{5.1} & 13.55 & 2.96 & 129.02 & \textbf{21.64} & 17.36 & \textbf{22.42} & \textbf{49} \\
DROID-VO~\cite{teed2021droid} & 98.43 & 84.2 & 108.8 & 2.58 & 0.93 & 59.27 & 64.4 & 24.2 & 64.55 & 71.8 & 16.91 & 54.19 & 17 \\
DPVO~\cite{teed2023deep} & 113.21 & 12.69 & 123.4 & 2.09 & \textbf{0.68} & 58.96 & 54.78 & 19.26 & 115.9 & 75.1 & 13.63 & 53.61 & 48\\
DROID-SLAM~\cite{teed2021droid} & 92.1 & 344.6 & X & {2.38} & 1.00 & 118.5 & 62.47 & 21.78 & 161.6 & X & 118.7 & - & 17\\
\midrule
DPV-SLAM & 112.8 & 11.50 & 123.53 & 2.50 & 0.81 & 57.80 & 54.86 & 18.77 & 110.49 & 76.66 & 13.65 & 53.03 & 39 \\

DPV-SLAM++ & 8.30 & 11.86 & 39.64 & {2.50} & 0.78 & {5.74} & \textbf{11.60} & \textbf{1.52} & 110.9 & {76.70} & {13.70} & {25.76} & 39 \\
\bottomrule
\end{tabular}
}
\caption{Comparison with other approaches using ATE[m].}\label{tab:1b}
\end{subtable}
\caption{Monocular SLAM on the KITTI~\cite{geiger2012we} training set. TartanVO only reports results for sequences 6,7,9,10 using the $t_{rel}/r_{rel}$ metrics. Compared to other general approaches to SLAM, our method performs well, while running at 39 FPS. \cite{deflowslam,goslam} do not report results on KITTI. This table shows the challenge of generalizing across domains: DROID-SLAM performs exceptionally well on indoor datasets, but struggles on KITTI. In turn, classical approaches perform well on KITTI, but are comparatively inaccurate indoors. DPV-SLAM++ performs well on both.}
\label{table:kitti}
\end{table}

Compared to other deep SLAM systems~\cite{teed2021droid, goslam, deflowslam}, our method performs similarly (0.054-0.076 vs 0.38-0.114), however these other approaches do not perform well (or don’t report results) in outdoor settings (see Tab. ~\ref{table:kitti}). Classical approaches generally do not perform well on this dataset. In contrast, they perform well on KITTI while previous deep SLAM methods do not. Our method performs well on both datasets.

\smallskip\noindent\textbf{KITTI~\cite{geiger2012we}} We evaluate monocular SLAM on sequences 00-10 from the KITTI training set in Tab.~\ref{table:kitti}. Video is recorded at 10FPS. The KITTI dataset includes long outdoor driving sequences with several loops. Scale drift is a known challenge in this setting. In order to correct scale drift, monocular methods must implement some form of loop closure. However, proximity-based loop detection is insufficient if there is significant scale drift, so none of the approaches relying solely on this mechanism perform well here. DPV-SLAM++, which also uses image retrieval, achieves the second-lowest average error among all reported methods, while averaging 39-FPS.%

\smallskip\noindent\textbf{EuRoC-MAV~\cite{burri2016euroc}} We evaluate monocular SLAM on the Machine-Hall and Vicon 1 \& 2 sequences from the EuRoC MAV dataset. Video is recorded at 20 FPS. This benchmark contains long sequences with motion blur, over/under-exposed images, and rapid camera movement. On EuRoC, our method performs similarly to DROID-SLAM (0.024 vs 0.022 ATE), while running 2.5x faster (50 FPS vs 20 FPS) using a quarter of the memory (5.0G vs 20G). Compared to the base DPVO system, we achieve 4x lower error (0.105$\rightarrow$0.024), with only a small reduction in speed (60$\rightarrow$50-FPS) and increase in cost (4G$\rightarrow$5G).

\begin{table} [t]
\centering
\resizebox{\linewidth}{!}{%
\begin{tabular}{cl| ccccc | ccc | ccc | c | c | c}
& & MH01 & MH02 & MH03 & MH04 & MH05 & V101 & V102 & V103 & V201 & V202 & V203 & Avg & FPS & VRAM\\
\toprule
& DeepFactors~\cite{czarnowski2020deepfactors} & 1.587 & 1.479 & 3.139 & 5.331 & 4.002 & 1.520 & 0.679 & 0.900 & 0.876 & 1.905 & 1.021 & 2.040 \\
& DeepV2D~\cite{teed2018deepv2d}$^\dagger$ & 0.739 & 1.144 & 0.752 & 1.492 & 1.567 & 0.981 & 0.801 & 1.570 & 0.290 & 2.202 & 2.743 & 1.298 \\
& DeepV2D [TartanAir]$^\dagger$ & 1.614 & 1.492 & 1.635 & 1.775 & 1.013 & 0.717 & 0.695 & 1.483 & 0.839 & 1.052 & 0.591 & 1.173 \\
& TartanVO$^1$~\cite{wang2021tartanvo}$^\dagger$ & 0.639 & 0.325 & 0.550 & 1.153 & 1.021 & 0.447 & 0.389 & 0.622 & 0.433 & 0.749 & 1.152 & 0.680 \\
& ORB-SLAM~\cite{orbslam1} & 0.071 & 0.067 & 0.071 & 0.082 & 0.060 & {\textbf{0.015}} & 0.020 & X & {0.021} & {0.018} & X & - \\
& DSO~\cite{gao2018ldso}$^\dagger$ & 0.046 & 0.046 & 0.172 & 3.810 & 0.110 & 0.089 & 0.107 & 0.903 & 0.044 & 0.132 & 1.152 & 0.601 \\
& LDSO~\cite{gao2018ldso} & 0.046 & 0.035 & 0.175 & 1.954 & 0.385 & 0.093 & 0.085 & - & 0.043 & 0.405 & - & - \\ %
& SVO~\cite{forster2014svo}$^\dagger$ & 0.100 & 0.120 & 0.410 & 0.430 & 0.300 & 0.070 & 0.210 & X & 0.110 & 0.110 & 1.080 & - \\
& ORB-SLAM3~\cite{orbslam3} & 0.016 & 0.027 & 0.028 & 0.138 & 0.072 & {0.033} & {0.015} & {0.033} & 0.023 & 0.029 & X & - \\
& DPVO~\cite{teed2023deep}$^\dagger$ & 0.087 & 0.055 & {0.158} & {0.137} & 0.114 & {0.050} & 0.140 & {0.086} & 0.057 & {0.049} & 0.211 & {0.105} \\
& GO-SLAM~\cite{goslam} & 0.016 & 0.014 & 0.023 & 0.045 & 0.045 & 0.037 & 0.011 & 0.023 & 0.016 & \textbf{0.010} & 0.022 & 0.024 & $6.8^\mathsection$ & 14G$^\mathsection$\\
& DROID-SLAM~\cite{teed2021droid} & \textbf{0.013} & \textbf{0.014} & {{0.022}} & {0.043} & 0.043 & 0.037 & 0.012 & 0.020 & \textbf{0.017} & 0.013 & \textbf{0.014} & \textbf{0.022} & 20 & 20G$^\mathsection$\\
\midrule
& {DPV-SLAM} & \textbf{0.013} & 0.016 & {0.022} & {0.043} & \textbf{0.041} & 0.035 & \textbf{0.008} & \textbf{0.015} & 0.020 & 0.011 & 0.040 & 0.024 & \textbf{50} & \textbf{5.0G}\\
& {DPV-SLAM++} & \textbf{0.013} & 0.016 & \textbf{0.021} & \textbf{0.041} & \textbf{0.041} & 0.035 & {0.010} & \textbf{0.015} & 0.021 & 0.011 & 0.023 & 0.023 & \textbf{50} & {7.0G}\\
\bottomrule
\end{tabular}
}
\caption{Monocular SLAM on the EuRoC datasets, ATE[m]. $^\dagger$ denotes visual odometry methods. We report runtimes and memory for methods whose average error is similar to ours. We perform marginally worse than DROID-SLAM (0.024 vs 0.022), but use significantly less GPU memory (5G vs 20G) and run 2.5x faster (50 FPS vs 20 FPS).}
\label{table:EurocMono}
\end{table}

\smallskip\noindent\textbf{TartanAir~\cite{wang2020tartanair}} We evaluate monocular SLAM on the TartanAir test set from the ECCV 2020 SLAM competition. Compared to existing approaches, DPV-SLAM outperforms DROID-SLAM by a sizeable margin (0.16 vs 0.24). Our method runs at 27 FPS, while DROID-SLAM runs at 8-FPS.

\begin{figure}[t]
\centering
\begin{subfigure}[b]{0.49\textwidth}
    \centering
    \includegraphics[width=\textwidth]{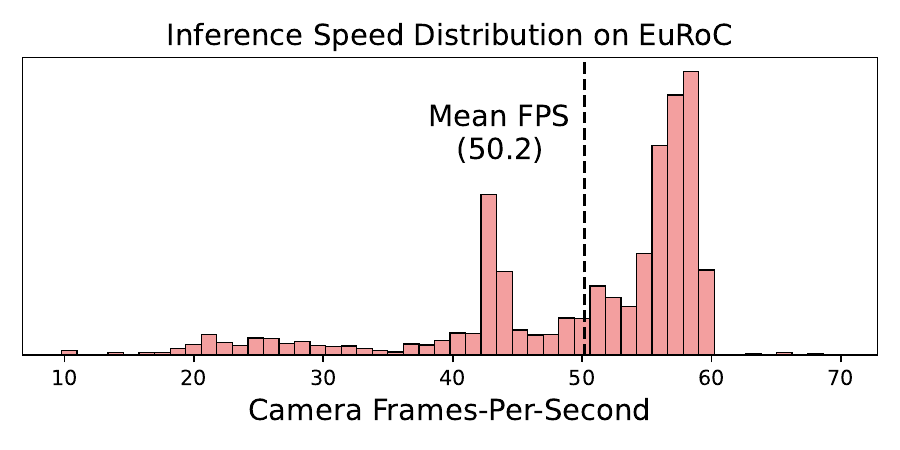} 
    \caption{Latency on EuRoC~\cite{burri2016euroc}}
\end{subfigure}
\begin{subfigure}[b]{0.49\textwidth}
    \centering
    \includegraphics[width=\textwidth]{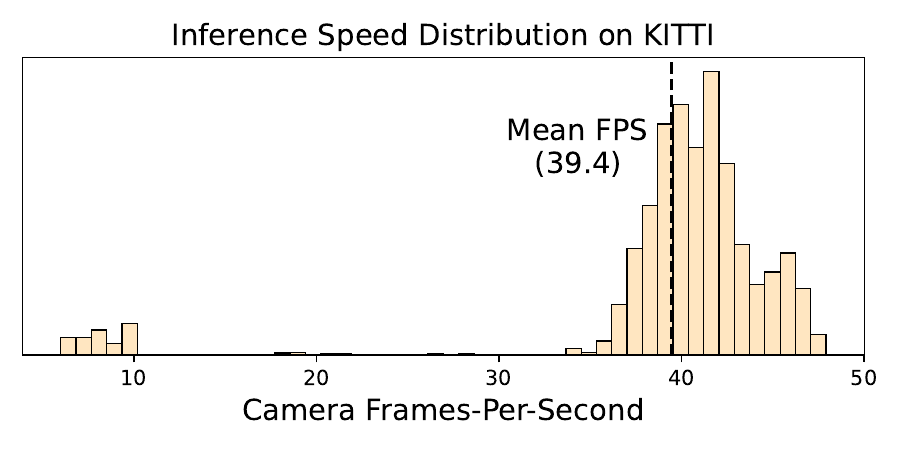} 
    \caption{Latency on KITTI~\cite{geiger2012we}}
\end{subfigure}
\caption{The distribution of inference speed. We run DPV-SLAM on EuRoC and KITTI, and sample the current FPS uniformly over the runtime. In both cases, there are two modes, the larger being representative of the odometry runtime and the smaller being representative of the loop closure. On EuRoC, the loop closure causes the runtime to drop from 58 to 42 FPS. On KITTI, the infrequent loop closure causes the speed to briefly dip below real-time (5-10 FPS) before returning to 40-FPS. The average speed on EuRoC and KITTI are 50-FPS and 39-FPS, or 2.5x and 3.9x real-time, respectively.}\label{fig:barplots}
\end{figure}

\begin{table}[h!]
\resizebox{\linewidth}{!}{%
\begin{tabular}{l|cccccccc | c}
Monocular & MH000 & MH001 & MH002 & MH003 & MH004 & MH005 & MH006 & MH007 & Avg \\
\toprule
ORB-SLAM~\cite{orbslam1} & 1.30 & \textbf{0.04} & 2.37 & 2.45 & X & X & 21.47 & 2.73 & - \\
DeepV2D~\cite{teed2018deepv2d} & 6.15 & 2.12 & 4.54 & 3.89 & 2.71 & 11.55 & 5.53 & 3.76 & 5.03 \\
TartanVO~\cite{wang2021tartanvo} & 4.88	& 0.26 & 2.00 & 0.94 & 1.07 & 3.19 & 1.00 & 2.04 & 1.92 \\
DPVO~\cite{teed2023deep} & 0.21 & \textbf{0.04} & {0.04} & 0.08 & 0.58 & {0.17} & 0.11 & 0.15 & {0.17} \\
DeFlowSLAM~\cite{deflowslam} & 0.63 & 0.06 & \textbf{0.02} & \textbf{0.01} & 2.80 & 0.20 & 0.31 & 0.45 & 0.56\\
DROID-SLAM~\cite{teed2021droid} & \textbf{0.08} & 0.05 & {0.04} & {0.02} & \textbf{0.01} & 1.31 & 0.30 & \textbf{0.07} & 0.24 \\
\midrule
DPV-SLAM & 0.23 & {0.05} & 0.04 & 0.04 & 0.54 & \textbf{0.15} & \textbf{0.07} & 0.14 & \textbf{0.16} \\
DPV-SLAM++ & 0.21 & \textbf{0.04} & 0.04 & 0.04 & 0.92 & {0.17} & 0.11 & 0.13 & 0.21 \\
\bottomrule
\end{tabular}
\caption{Results on the TartanAir monocular benchmark (ATE[m]). We outperform existing approaches.}
\label{table:TartanAir}
}
\end{table}
\newpage
\section{Limitations} 

DPV-SLAM requires a GPU, as opposed to classical approaches, and only provides a sparse 3D reconstruction. The cost of the global bundle adjustment layer also scales quadratically with the number of pose variables, which is why we limit its range to 1000 (key) frames. See the Appendix for details.

DPV-SLAM, like most monocular systems, sometimes suffers from scale-drift in outdoor environments, though orthogonal work has addressed this problem using monocular depth networks~\cite{yin2023metric3d}. The image retrieval is also susceptible to the occasional false-positive detection. The classical loop closure also incurs an additional 2G GPU memory overhead due to the invocation of the U-Net keypoint detector~\cite{tyszkiewicz2020disk}.

\section{Conclusion} 
We introduce DPV-SLAM, a system for monocular visual SLAM. DPV-SLAM generalizes well to different domains, and is efficient. It runs at a relatively consistent frame-rate and only requires a single GPU. We evaluate on EuRoC, TartanAir, TUM-RGBD and KITTI. We show that our approach is robust across domains, and is comparable to, or better than the SOTA on several datasets, while running faster and using less compute. We hope this will be a valuable resource for the computer vision community. This work was partially supported by the National Science Foundation.

\bibliographystyle{splncs04}
\bibliography{main}

\appendix

\section{Bundle-Adjustment Efficiency}
\begin{figure}[h]
\centering
\begin{subfigure}[b]{0.49\textwidth}
    \centering
    \includegraphics[width=\textwidth]{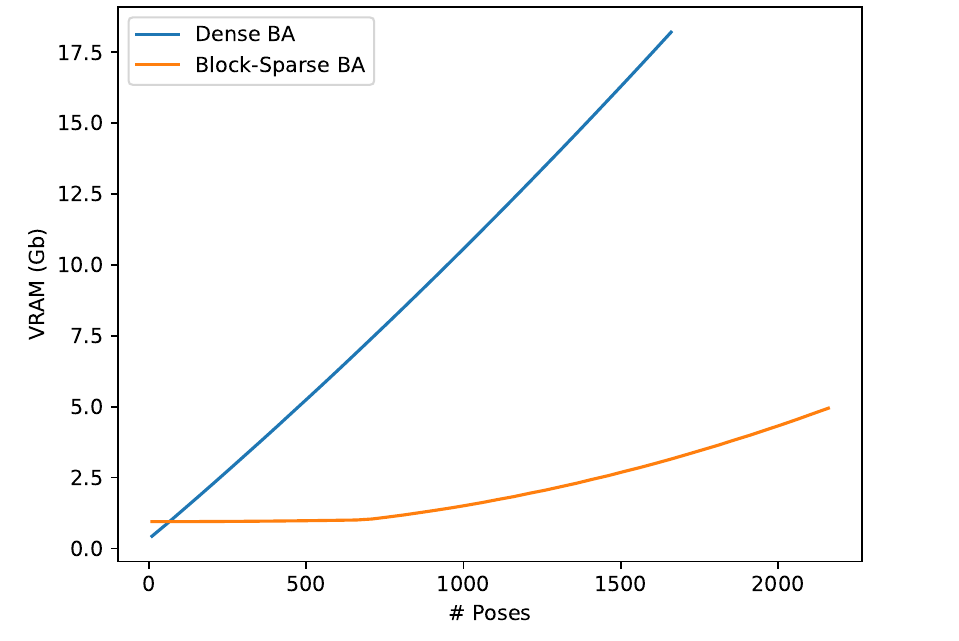} 
    \caption{BA VRAM on KITTI.}
\end{subfigure}
\begin{subfigure}[b]{0.49\textwidth}
    \centering
    \includegraphics[width=\textwidth]{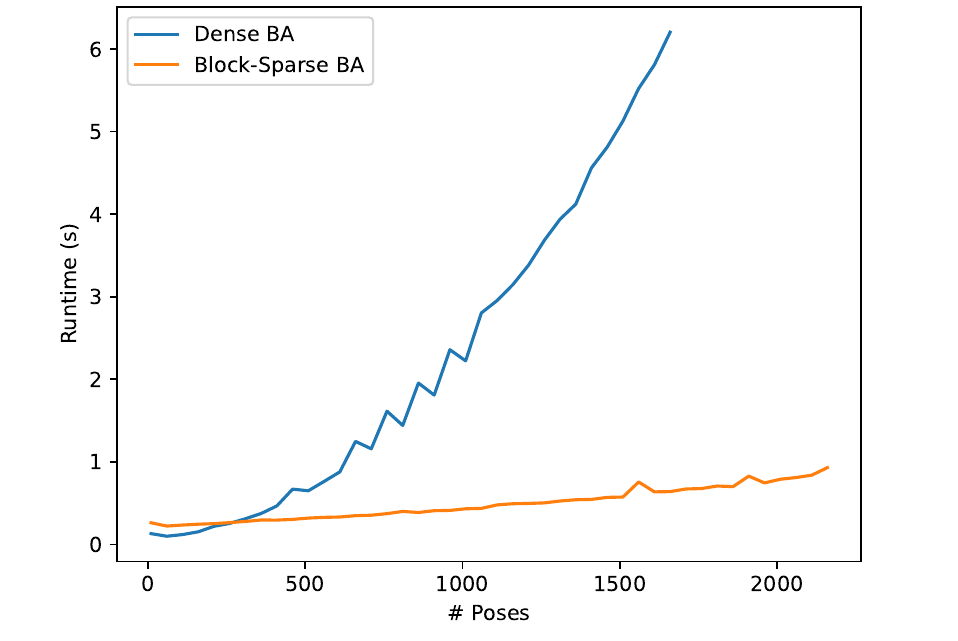} 
    \caption{BA runtime on KITTI.}
\end{subfigure}
\caption{The cost of running global optimization on patch graphs, comparing our block-sparse implementation to the dense one. The former is significantly faster and cheaper for large patch graphs, but slightly less optimal for small graphs. We switch between the two implementations on-the-fly depending on the size.}
\label{fig:bacost}
\end{figure}

In Fig.~\ref{fig:bacost}, we show the cost of bundle adjustment on the KITTI dataset as a function of the number of pose variables. In this experiment, we lift the 1000-pose limit which we used to obtain results in the main paper. We compare the dense implementation from DPVO with our block-sparse implementation. As expected, the block-sparse implementation is much more efficient for moderate-to-large patch graphs. For very small graphs (i.e., odometry-only), the dense implementation is better since the patch graph has high connectivity and lacks the additional indexing overhead from the block-sparse version. The number of depth variables and edges remains constant in these experiments. In the block sparse version, the Cholesky decomposition of the pose-block in the schur complement in t is the memory and runtime bottleneck.

\end{document}